USING ENTITY RELATIONS FOR OPINION MINING OF VIETNAMESE COMMENTS

Nguyen Tan Phat[1], Le Tan Loi[1], Ngo Minh Vuong[2,*], Nguyen Minh Phuc[2]

[1]*Faculty of Computer Science and Engineering, HoChiMinh City University of Technology, 268 Ly Thuong Ket, District 10, Ho Chi Minh City*

[2]*Faculty of Information Technology, Lac Hong University, 10 Huynh Van Nghe, Bien Hoa City, Dong Nai Province*

[*]Email: *ngovuong@lhu.edu.vn*

**Abstract:** In this paper, we propose several novel techniques to extract and mining opinions of Vietnamese reviews of customers about a number of products traded on e-commerce in Vietnam. The assessment is based on the emotional level of customers on a specific product such as mobile and laptop. We exploit the features of the products because they are much interested by customers and have many products in the Vietnam e-commerce market. Thence, it can be known the favorites and dislikes of customers about exploited products.

*Keywords*: opinion mining, sentiment analysis, ontology, named entity.

# SỬ DỤNG MỐI QUAN HỆ THỰC THỂ CHO KHAI THÁC Ý KIẾN CỦA CÁC BÌNH LUẬN TIẾNG VIỆT

**Nguyễn Tấn Phát[1], Lê Tấn Lợi[1], Ngô Minh Vương[2,*], Nguyễn Minh Phúc[2]**

[1]*Khoa KH&KT Máy Tính, Trường Đại Học Bách Khoa TpHCM, 268 Lý Thường Kiệt, Quận 10, Tp. Hồ Chí Minh*

[2]*Khoa Công Nghệ Thông Tin, Trường Đại Học Lạc Hồng, 10 Huỳnh Văn Nghệ (Tỉnh Lộ 24), Tp. Biên Hòa, Tỉnh Đồng Nai*

*Email: ngovuong@lhu.edu.vn

## TÓM TẮT

Trong công trình này chúng tôi đề xuất một phương pháp để rút trích và khai thác những ý kiến đánh giá tiếng Việt của khách hàng về các sản phẩm được bán trên các trang web thương mại điện tử ở Việt Nam. Việc đánh giá này dựa trên mức độ thể hiện tình cảm của khách hàng trên một sản phẩm cụ thể như điện thoại di động và máy tính xách tay. Chúng tôi chọn những sản phẩm này vì chúng có số lượng lớn khách hàng quan tâm, cũng như có số lượng phong phú và đa dạng có trên thị trường thương mại điện tử ở Việt Nam. Từ đó có thể biết được sự yêu thích hoặc chưa hài lòng của khách hàng về các sản phẩm cần khảo sát.

*Từ khóa:* khai thác ý kiến, phân tích tình cảm, ontology, thực thể có tên.

## 1. GIỚI THIỆU

Trong thời đại này nay, có rất nhiều người sử dụng và chia sẻ thông tin trên mạng Internet. Ở đó, những thông tin được trao đổi hàng ngày và có rất nhiều ý kiến, bình luận của mọi người được chia sẻ. Vì thế việc xem xét và hiểu được các nhận xét này là rất hữu ích. Hiện nay có nhiều công trình tìm hiểu về các ý kiến trên các tạp chí online ([1]). Ngoài ra, Mạng xã hội như Twitter và Facebook phù hợp cho nhiều ứng dụng phân tích tâm lý, quan điểm của mọi người ([2]). Ví dụ như giám sát danh tiếng của một thương hiệu cụ thể trên mạng xã hội, cung cấp cho các ứng viên tham gia ứng cử biết được nguyện vọng của cử tri, từ đó điều chỉnh phát biểu và hành động. Hệ thống phân tích nhận xét cũng có thể ứng dụng vào việc phân tích thị trường tài chính như thị trường chứng khoán.

Trong các mạng xã hội, Twitter đang trở thành nguồn cung cấp tin chính vì Twitter là một dạng siêu blog với hơn 340 triệu tweet (dạng tin nhắn ngắn) mỗi ngày. Vì thế việc khai thác ý kiến trên Twitter cũng đang được khai thác ([3], [4]). Với mỗi ngôn ngữ sẽ có một đặc trưng khác nhau nên việc phân tích ý kiến trên các ngôn ngữ khác nhau sẽ cần đến các kỹ thuật khác nhau. Các công trình phân tích ý kiến phần lớn tập trung vào tiếng Anh. Có một số ít công trình nghiên cứu nhận xét ở tiếng Ả Rập ([5], [6]). Công trình [7] có nghiên cứu về các nhận xét tiếng Việt nhưng nhằm mục đích xem các nhận xét này có là nhận xét rác hay không. Trong khi đó, công trình của chúng tôi là phân tích ý kiến của các nhận xét tiếng Việt.

Hầu hết các ứng dụng của phân tích ý kiến là nằm trong nhận xét của khách hàng về sản phẩm và dịch vụ ([2]). Vì thế trong công trình này, chúng tôi sẽ khảo sát các ý kiến trên các website thương mại điện tử. Thương mại điện tử là một bước đột phá so với kinh doanh theo truyền thống (người bán và người mua trao đổi trực tiếp với nhau) và đang trở nên rất phổ biến. Số lượng khách hàng sử dụng dịch vụ mua bán qua mạng càng nhiều và sự quan tâm của họ với loại hình kinh doanh này cũng ngày càng tăng lên. Do đó, sẽ có rất nhiều thậm chí hàng trăm, hàng nghìn những lời bình luận, nhận xét cho những sản phẩm hoặc dịch vụ mà họ quan tâm hoặc những sản phẩm, dịch vụ đang phổ biến trên thị trường như máy ảnh, điện thoại di động, máy tính xách tay, phim điện ảnh chiếu rạp, sách, khách sạn và du lịch. Chính vì vậy, thật khó để cho những khách hàng tiềm năng có thể tìm đọc hết những lời bình luận, nhận xét của những khách hàng trước đó đã sử dụng để có thể đưa ra được những quyết định hợp lý. Và cũng thật khó để các nhà sản xuất sản phẩm đó có thể theo dõi và quản lý những ý kiến của khách hàng để làm thỏa mãn khách hàng.

Chính vì những lý do trên, hệ thống phân tích quan điểm của nhận xét tiếng Việt của chúng tôi nhằm mục đích khai thác và tổng hợp lại những bày tỏ ý kiến, những bình luận về những sản phẩm đang thịnh hành trên các trang web thương mại điện tử như điện thoại di động và máy tính xách tay. Hệ thống của chúng tôi sẽ xác định chủ đề nào được đề cập đến trong nhận xét. Một nhận xét có thể có nhiều hơn một thực thể được nói đến, ví dụ như trong nhận xét kiểu so sánh. Một thực thể có thể được đề cập đến thông qua các đặc trưng, thuộc tính con của nó. Tiếp theo hệ thống sẽ tìm kiếm các ý nghĩa mang tính tích cực, tiêu cực hoặc trung tính trong nhận xét. Từ đó hệ thống sẽ xác định được quan điểm của người viết nhận xét cho chủ đề được đề cập trong nhận xét này.

## 2. XÁC ĐỊNH NGỮ NGHĨA CỦA NHẬN XÉT THÔNG QUA ONTOLOGY VÀ TẬP TỪ ĐIỂN

### 2.1 Ontology

Ontology bắt nguồn từ triết học, được dẫn xuất từ tiếng Hy Lạp là "onto" và "logia". Trong những năm gần đây, ontology được sử dụng nhiều trong khoa học máy tính và được định nghĩa khác với nghĩa ban đầu. Theo đó ontology là sự mô hình hóa và đặc tả các các khái niệm một cách hình thức, rõ ràng và chia sẻ được ([8], [9]). Ngoài ra, ontology cần có thêm tính thống nhất, tính mở rộng và tính suy luận. Trong mô hình này chúng tôi nghiên cứu về 2 miền đó là về điện thoại di động và máy tính xách tay. Hai miền này sẽ mô tả thông qua các khái niệm của các thực thể dữ liệu được quan tâm, các thuộc tính, và các mối liên hệ. Các khái niệm này được tổ chức theo cấu trúc cây phân cấp. Mỗi nút đại diện cho một khái niệm, và mỗi khái niệm là một chuyên biệt hoá của nút cha.

Hình 1 thể hiện hai miền Ontology là điện thoại di động và máy tính xách tay. Hai miền này chúng tôi thu thập được các thực thể tương ứng như điện thoại, máy tính Samsung, Apple, iPhone, Nokia và Asus. Để nhận biết được các thành phần thuộc tính của chủ đề trên thì chúng tôi đã thu thập các thuộc tính chung của hai dòng sản phẩm máy tính xách tay và điện thoại di động như cảm ứng, hệ điều hành, camera, flash, giá cả và bàn phím.

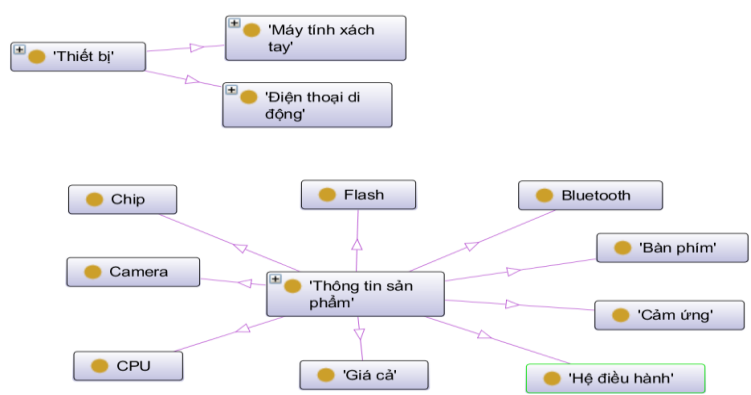

*Hình 1*. Miền Ontology về máy tính xách tay và điện thoại di động.

Trong một nhận xét bao gồm các quan điểm về những thành phần khác nhau xoay quanh chủ đề được đề cập. Vì vậy việc xác định các từ hay cụm từ liên quan đến chủ đề cũng như mô tả những vấn đề con của chủ đề là một bước then chốt trong quá trình xử lý. Với mối quan hệ của các thành phần trong Ontology, chúng tôi có thể dễ dàng xác định được các thành phần con của chủ đề chính. Nhờ vậy chúng tôi thuận lợi hơn trong việc xác định được các chủ đề con mà tác giả nhắc đến trong nhận xét. Bên cạnh đó, kinh nghiệm cho thấy những từ ngữ hay cụm từ thể hiện những vấn đề liên quan đến chủ đề thường là các danh từ, hay đại từ, vì vậy để có thể chính xác và nhanh chóng hơn trong việc phát hiện và xác định các từ hay cụm từ này, chúng tôi cần gán nhãn cho các thành phần trong câu trước khi xử lý.

Ví dụ: *"Ipad mini có dung lượng pin thấp. Cứ Nexus mà xài cho lành, giá rẻ, cấu hình cũng ngon."*

Ở ví dụ trên việc xác định các thực thể có tên như Ipad mini, Nexus và các thành phần, thuộc tính của những sản phẩm này như dung lượng pin, giá, cấu hình sẽ dựa vào Ontology. Dùng Ontology để xác định đúng những thuộc tính tương ứng với từng chủ đề cụ thể. Từ đó chúng tôi sẽ xác định được ngữ nghĩa trong một nhận xét của người dùng. Thực thể có tên là con người, tổ chức, nơi chốn, và những đối tượng khác được tham khảo bằng tên ([10]).

### 2.2. Tập từ điển

*Bảng 1*. Ví dụ một số động từ/tính từ có thể hiện tình cảm.

| Những từ thể hiện tình cảm | Phân loại |
|---|---|
| Buồn cười, buồn buồn, chán, chậm chạp | Động từ (-) |
| Cao cấp, đáng kể, đàng hoàng | Tính từ (+) |
| Xấu, sến, thô kệch, tẻ nhạt | Tính từ (-) |
| Đặc biệt, đặc sắc, hết sẩy, hết ý, yêu, thích | Động từ (+) |

Để có thể tính toán được mức độ tích cực của mỗi quan điểm sau khi xác định được các từ ngữ thể hiện quan điểm, trước tiên chúng tôi cần biết được sự phân cực của các từ ngữ này là tiêu cực (-) hay tích cực (+). Vì vậy, tập từ điển (B) trong Hình 2 là một phần của hệ thống, đây là tập hợp các tính từ và động từ (thể hiện quan điểm), đồng thời kèm theo sự phân cực của chúng. Với

Ontology này, chúng tôi sẽ có được dữ liệu cho quá trình tính toán độ tích cực của một quan điểm. Ví dụ: Đẹp (+), xấu (-), thích (+), ghét (-).

Bảng 1 liệt kệ một số ví dụ về tính từ và động từ có thể hiện tình cảm. Trong đó, những tính từ và động từ thể hiện tính tích cực như đặc biệt, đặc sắc, cao cấp, sang trọng ,và những tính từ và động từ thể hiện tính tiêu cực như chán, chậm chạp, xấu, thô kệch.

### 3. XỬ LÝ VÀ TÍNH TOÁN MỨC ĐỘ THỂ HIỆN TÌNH CẢM CÓ TRONG NHẬN XÉT

#### 3.1. Kiến trúc hệ thống

Hệ thống bao gồm bảy module, một tập Ontology và một tập từ điển được thể hiện như Hình 2

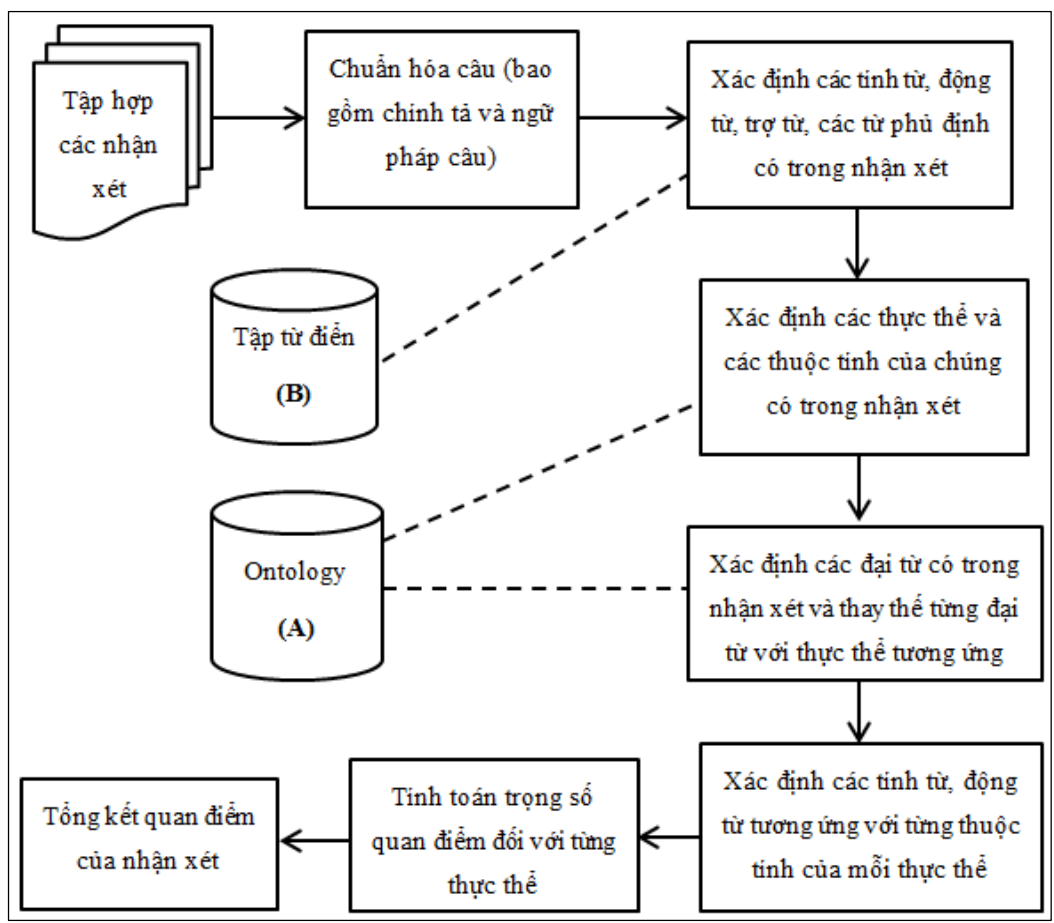

*Hình 2*. Mô hình đánh giá các nhận xét dựa trên tập từ điển và Ontology.

Vấn đề nhận xét một quan điểm, đó là một vấn đề liên quan nhiều đến xử lý ngôn ngữ. Để nhận định một nhận xét có tính tích cực, tiêu cực hoặc trung tính, chúng tôi cần xác định nhiều yếu tố, một trong những yếu tố then chốt của vấn đề là xác định những từ (cụm từ) thể hiện quan điểm. Thêm vào đó, hiệu suất của hệ thống sẽ được gia tăng nếu xem xét đến chủ đề của nhận xét. Ví dụ như chúng tôi càng chi tiết về tỉ lệ những nhận xét tốt hay xấu cho một sản phẩm ở mức độ các thành phần của sản phẩm thì nhà sản xuất sẽ có nhiều thông tin hơn từ khách hàng cho từng yếu tố của sản phẩm và đưa ra biện pháp khắc phục tốt hơn.

Ngoài ra, những nhận xét thực tế chứa đựng rất nhiều lỗi phổ biến như chính tả và ngữ pháp. Những yếu tố này ảnh hưởng nhiều về độ sai lệch giữa kết quả hệ thống so với thực tế, ví như chúng tôi bỏ qua các nhận định có lỗi chính tả hay tưởng nhầm đó là một câu bình thường.

Vì vậy, hệ thống đạt được kết quả chính xác cần tích hợp được các tiến trình xử lý các vấn đề nêu trên. Kiến trúc đề xuất đã thể hiện những vấn đề được nhìn nhận và đưa ra một phương pháp giải quyết tổng quát. Hệ thống được xây dựng trên kiến trúc đề xuất, là một tiến trình tổng quát, bao gồm nhiều tiến trình con, những tiến trình đi trước tạo ra kết quả, các kết quả này qua từng tiến trình dần được tinh chế, để chỉ chứa các yếu tố cần thiết cho kết quả cuối cùng, và tiến trình sau sẽ nhận kết quả của tiến trình trước để xử lý tiếp.

Kiến trúc đã giải quyết được những yếu tố ảnh hưởng đến kết quả của hệ thống và thể hiện được tính kết nối giữa các module, đó là tính tổng quan của kiến trúc, và hệ quả là một kết quả chính xác sau khi thực hiện tốt các module đề ra.

Như đã đề cập, việc xác định các từ (cụm từ) liên quan đến chủ đề là một tiến trình quan trọng của hệ thống, giúp tăng độ chính xác và nâng cao giá trị của hệ thống. Ontology có cấu trúc phân lớp danh từ và thể hiện mối quan hệ giữa chúng với nhau, với những đặc tính đó, Ontology giúp chúng tôi giải quyết dễ dàng và chính xác nhất cho quá trình xác định chủ đề mà nhận xét hướng đến.

## 3.2. Đánh giá điểm số

Chúng tôi tiến hành đánh giá nhận xét thông qua việc cho điểm các từ ngữ mang tính sắc thái tình cảm tồn tại trong nhận xét. Theo cách này chúng tôi sẽ thực hiện việc đánh giá điểm số cho từng tính từ và động từ cụ thể. Kết quả điểm số của một nhận xét là giá trị đại số của điểm số các tính từ và động từ có liên quan trong nhận xét.

Trong nhận xét có thể có nhiều câu, với mỗi câu chúng ta xác định danh sách tính từ, danh sách động từ cho từng thực thể. Từ đó hệ thống xác định được danh sách tính từ, động từ bổ nghĩa cho từng thực thể trong toàn nhận xét chính là hợp của các danh sách tính từ, động từ bổ nghĩa cho thực thể tương ứng trong từng câu.

Vấn đề trọng yếu là xác định được danh sách tính từ động từ cho từng thực thể trong câu. Vấn đề này được giải quyết bằng cách thực hiện việc xác định các thuộc tính của thực thể tương ứng. Ứng với từng thuộc tính, sẽ có một từ hay cụm từ bổ nghĩa cho nó, được xác định thông qua vị trí tương đối so với thuộc tính. Các từ, cụm từ bổ nghĩa thường đứng trước thuộc tính và chủ đề được đề cập.

Sau khi đã có được danh sách tính từ, động từ bổ nghĩa cho từng thực thể (chủ đề), chúng ta sử dụng các công thức sau để tính toán trọng số quan điểm của nhận xét trên từng thực thể (chủ đề).

Công thức:

$$score(entity) = \frac{\sum_{i=0}^{size-1} score(adjList[i]) + \sum_{i=0}^{size-1} score(verbList[i])}{\sum_{i=0}^{size-1} |score(adjList[i])| + \sum_{i=0}^{size-1} |score(verbList[i])|} \qquad (1)$$

$$score(adj) = s(adj)*s(auWord)*specialCase*negative$$

$$score(verb) = s(verb)*s(auWord)*negative$$

Trong đó:

- score(entity): là điểm số thể hiện quan điểm của nhận xét với thực thể, thuộc đoạn [-1, 1].
- adjList: là danh sách tính từ bổ nghĩa cho thực thể tương ứng. Việc xác định danh sách các tính từ này dựa vào tập từ điển.
- verbList: là danh sách động từ bổ nghĩa cho thực thể tương ứng. Việc xác định danh sách các động từ này dựa vào tập từ điển.
- size: kích thước của danh sách tương ứng (danh sách tính từ hay động từ). Kích thước là số lượng các tính từ hoặc động từ có trong nhận xét.
- score(adj): điểm số của tính từ ảnh hưởng đến quan điểm trong nhận xét.
- score(verb): điểm số của động từ ảnh hưởng đến quan điểm trong nhận xét.
- s(adj): trọng số của tính từ, thuộc đoạn [-1,1]. Trọng số của tính từ được ánh xạ từ tập từ điển với tính từ tương ứng.
- s(verb): trọng số của động từ, thuộc đoạn [-1,1]. Trọng số của động từ được ánh xạ từ tập từ điển với động từ tương ứng.
- s(auWord): trọng số của trợ từ bổ nghĩa cho động từ (tính từ) tương ứng, thuộc đoạn [-1.5; 1.5]. Trọng số của trợ từ được ánh xạ từ tập từ điển với trợ từ tương ứng.
- negative: thể hiện sự xuất hiện của từ phủ định, ảnh hưởng đến cực của quan điểm, thuộc tập {-1; 1}. Các từ phủ định có trong câu được xác định thông qua tập từ điển.
- specialCase: đây là trường hợp thể hiện mối quan hệ đặc biệt của tính từ và thực thể mà nó bổ nghĩa, làm thay đổi cực quan điểm của tính từ, specialCase thuộc tập {-1; 1}. Việc xác định specialCase dựa vào tập Ontology về mối quan hệ giữa tính từ và thực thể.

Một số vấn đề về việc đánh giá điểm số cho từng thực thể được chúng tôi thể hiện như sau:

a) Đối với loại nhận xét chỉ có một câu và có một thực thể:

Chúng tôi cần xác định thực thể đang được nhắc đến dựa vào Ontology. Sau đó chúng ta xác định danh sách tính từ và động từ bổ nghĩa cho thực thể đó.

Dựa vào danh sách tính từ và động từ ta xác định điểm số thể hiện quan điểm cho thực thể như công thức (1).

b) Đối với loại nhận xét có một câu và có nhiều thực thể:

Chúng tôi tiến hành xác định danh sách thực thể trong câu dựa vào Ontology. Sau đó chúng tôi xác định danh sách tính từ và động từ bổ nghĩa cho từng thực thể. Danh sách tính từ động từ bổ nghĩa cho từng thực thể gồm các tính từ, động từ đứng ngay sau thực thể tương ứng và đứng ngay trước thực thể kế tiếp (nếu tồn tại).

Dựa vào danh sách tính từ, động từ bổ nghĩa cho từng thực thể, áp dụng công thức (1), chúng tôi xác định được trọng số thể hiện quan điểm của nhận xét cho thực thể tương ứng.

c) Đối với loại nhận xét có nhiều câu và chỉ có một thực thể:

Chúng tôi tiến hành xác định thực thể mà nhận xét nhắc đến dựa vào Ontology. Tiếp theo chúng tôi xác định danh sách các tính từ và động từ bổ nghĩa cho thực thể. Danh sách tính từ và động từ này bao gồm các tính từ và động từ đứng ngay sau thực thể trong phạm vi toàn nhận xét.

Dựa vào danh sách tính từ và động từ, áp dụng công thức (1), chúng tôi xác định được điểm số thể hiện quan điểm của nhận xét đối với thực thể mà nhận xét đã đề cập.

d) Đối với loại nhận xét có nhiều câu và có nhiều thực thể:

Chúng tôi tiến hành xác định danh sách thực thể mà nhận xét nhắc đến dựa vào Ontology trong tất cả các câu, bằng cách ứng với mỗi câu hệ thống xác định các thực thể trong câu. Ứng với mỗi thực thể, chúng tôi sẽ xác định các vị trí mà nó xuất hiện trong từng câu. Dựa vào vị trí của thực thể trong từng câu, chúng ta sẽ xác định được danh sách tính từ và động từ thể hiện quan điểm bổ nghĩa cho nó.

Dựa vào danh sách tính từ và động từ, áp dụng công thức (1), hệ thống xác định được điểm số thể hiện quan điểm của nhận xét đối với từng thực thể tương ứng, *score(enity)>0*: tích cực, *score(enity)<0:* tiêu cực, *score(enity)=0:* trung tính.

## 4. THỰC NGHIỆM

Trước tiên chúng tôi xin giới thiệu về tập thí nghiệm được sử dụng để kiểm tra hệ thống đánh giá. Chúng tôi có tất cả ba tập kiểm tra, tập dataset 1 chúng tôi thu thập gồm 200 câu bình luận của khách hàng trên các trang web thương mại điện tử, các diễn đàn về công nghệ thông tin[1]. Ở tập dataset 1 này, mỗi bình luận chỉ nói về một thực thể và chủ yếu nhận xét các dòng điện thoại di động. Tập dataset 2 chúng tôi thu thập gồm 100 câu bình luận về chủ đề điện thoại di động và máy tính xách tay, mỗi bình luận nhận xét về nhiều thực thể. Và tập dataset 3 chúng tôi thu thập gồm 100 câu bình luận trong đó kết hợp các bình luận về một thực thể và nhiều thực thể với nhau về chủ đề điện thoại di động và máy tính xách tay.

Tiếp theo, chúng tôi sẽ giới thiệu về tập huấn luyện về đánh giá về điểm số và kết quả thí nghiệm trên tập dữ liệu gồm 100 câu bình luận mà chúng tôi đã thực hiện được. Về tập huấn luyện, chúng tôi thu thập 100 câu bình luận về nhiều thực thể, trong đó các thực thể thuộc chủ đề về điện thoại di động và máy tính xách tay. Về kết quả thực nghiệm, chúng tôi tiến hành thực nghiệm trên ba tập dữ liệu được chúng tôi giới thiệu như trên.

### 4.1. Tập huấn luyện

Về cách đánh giá điểm số cuối cùng mà hệ thống thực hiện được như trên, hệ thống sẽ tiến hành chuẩn hóa điểm số trong đoạn [-1;1] và cuối cùng đưa ra kết luận là tích cực(positive), tiêu cực(negative) và trung tính(neutral). Việc xác định điểm số để hệ thống nhận biết một nhận xét về thực thể trong bình luận là trung tính hay không là một trường hợp nhạy cảm có khi con người cũng không thể nhận biết được. Vì thế chúng tôi đã tiến hành thử nghiệm trên tập huấn luyện này để rút ra điểm số đáng tin cậy nhất cho việc xác định điểm số trung tính cho thực thể.

Về cách hiện thực, chúng tôi cho chương trình chạy tự động điểm số trong đoạn [-1;1], giữ giá trị cận trên upper và cho giá trị cận dưới under chạy, mỗi lần giá trị cận dưới giảm xuống 0.05. Nếu kết quả trung bình của P và R lần sau mà nhỏ hơn lần trước đó thì chúng tôi không ghi vào kết quả thực hiện như Bảng 2.

---

[1] www.thegioididong.com, www.vatgia.com, www.tinhte.vn

*Bảng 2.* Đánh giá điểm số trung tính

| Upper | Under | P (%) | R (%) |
|---|---|---|---|
| 1,0 | -1,0 | 25,03 | 30,00 |
| 1,0 | -0,95 | 31,69 | 35.67 |
| 1,0 | -0,65 | 32,03 | 36,00 |
| 1,0 | -0,5 | 32,36 | 36,33 |
| 1,0 | -0,3 | 32,69 | 36,67 |
| 1,0 | -0,2 | 33,36 | 37,33 |
| 0,95 | -1,0 | 48,56 | 50,39 |
| 0,95 | -0,95 | 60,39 | 61,39 |
| 0,95 | -0,65 | 61,06 | 62,06 |
| 0,95 | -0,5 | 61,72 | 62,72 |
| 0,95 | -0,3 | 62,39 | 63,39 |
| 0,95 | -0,2 | 63,72 | 64,72 |
| 0,65 | -0,2 | 63,89 | 65,06 |
| 0,5 | -0,2 | 65,22 | 66,83 |
| 0,3 | -0,2 | 66,00 | 67,83 |
| 0,1 | -0,2 | 66,22 | 68,61 |
| 0,1 | -0,1 | 66,56 | 68,94 |
| 0 | 0 | 67,44 | 70,22 |

Ở Bảng 2 là kết quả chúng tôi đã tiến hành thực hiện được trên tập huấn luyện. Như ta thấy, trên cột Upper là giá trị điểm số trên thuộc đoạn điểm số chuẩn hóa, cột Under là giá trị điểm số dưới thuộc đoạn điểm số chuẩn hóa và hai cột tiếp theo tương ứng là độ chính xác (P) và độ đầy đủ (R) trên mỗi mốc điểm số. Ứng với mỗi mốc điểm số sẽ giảm giá trị Upper xuống 0,05 và tăng giá trị Under lên 0,05. Hệ thống sẽ tiến hành phân tích các nhận xét trên tập huấn luyện với điểm số trung tính thuộc đoạn [Upper; Under]. Ở mỗi mốc điểm số, sau khi hệ thống đánh giá sẽ in ra kết quả nếu độ chính xác của mốc điểm số sau lớn hơn mốc điểm số trước, ngược lại hệ thống sẽ bỏ qua kết quả với mốc điểm số này. Như vậy, từ Bảng 2 chúng tôi rút ra được kết luận là lấy điểm 0 làm điểm đánh giá mức độ trung tính của một thực thể với độ chính xác cao nhất là 67,44 %.

## 4.2. Kết quả thí nghiệm

Sau khi tiến hành thực hiện thí nghiệm trên tập huấn luyện để xác định được điểm số thích hợp cho việc đánh giá mức độ trung tính của một thực thể, chúng tôi tiếp tục tiến hành thử nghiệm trên tập 200 câu bình luận về một thực thể ở mỗi bình luận, 100 câu bình luận về nhiều thực thể ở mỗi bình luận và 100 câu bình luận kết hợp về một thực thể lẫn nhiều thực thể ở mỗi bình luận.

*Bảng 3.* Đánh giá kết quả thí nghiệm

| Tập dữ liệu | Số câu | Tích cực (%) | | Tiêu cực (%) | | Trung tính (%) | | **Trung bình (%)** | |
|---|---|---|---|---|---|---|---|---|---|
| | | Độ P | Độ R | Độ P | Độ R | Độ P | Độ R | **Độ P** | **Độ R** |
| Dataset 1 | 200 | 90 | 97,3 | 91,36 | 94,87 | 89,74 | 72,92 | **90,37** | **88,36** |
| Dataset 2 | 100 | 69 | 76 | 70,33 | 72 | 63 | 62,67 | **67,44** | **70,22** |
| Dataset 3 | 100 | 82 | 82 | 87,5 | 88 | 84 | 86 | **84,5** | **85,33** |

Ở Bảng 3 thể hiện kết quả của ba tập dữ liệu, kết quả cho ta thấy độ chính xác trên tập bình luận về một thực thể là 90,37 %, trên tập bình luận về nhiều thực thể là 67,44 % và trên tập bình luận về một hoặc nhiều thực thể là 84,5 %. Qua kết quả trên cho ta biết việc đánh giá bình luận trên một thực thể đơn giản hơn việc đánh giá bình luận trên nhiều thực thể. Bởi vì đối với bình luận nhiều thực thể có nhiều thực thể được nói đến nên hệ thống khó xác định được người sử dụng đang đề cập đến thực thể nào. Việc so sánh giữa nhiều thực thể với nhau cũng làm cho hệ thống đánh giá nhầm lẫn giữa các thực thể này. Trong khi đó, khi hệ thống đánh giá trên bình luận một thực thể thì hệ thống dễ xác định được những tính từ, động từ nào đang nói về thực thể trong câu và hệ thống sẽ xác định được điểm số chính xác nhất.

Ví dụ 1: *"máy này xài ngon, tui vừa mua về xài được 1 tuần, đã quen thuộc với máy, pin ổn, chức năng tuyệt vời!!!"*

Ví dụ 2: *"Nhìn không ưng mắt mất đi vẻ sang trọng của nó"*

Ví dụ 3: *"Tính năng ổn, tuy nhiên giá hơi cao!"*

Ví dụ 4: *"Nexus hiện giờ pin rất kém. Chỉ mong bản sau pin tốt hơn là mua ngay."*

Ví dụ 5: *"xem ra con Asus này cũng không hay cho lắm, mua dell vostro 3450 có hơn 0,2 kg xem ra còn hay hơn."*

Trong những ví dụ (vd) trên, từ vd 1 đến vd 4 là những bình luận về một thực thể. Trong đó vd 1, vd 2 và vd 3 là những ví dụ mà hệ thống đánh giá đúng tương ứng tích cực, tiêu cực, và trung tính. Hệ thống đánh giá đúng nhờ vào sự phân tích giữa các từ mang ý nghĩa thể hiện tình cảm như ngon, ổn, tuyệt vời (ở vd 1) với các thuộc tính của một sản phẩm như pin, chức năng mang ý nghĩa tích cực. Trong vd 2 hệ thống cũng xác định được những phủ định từ như không, mất, những phủ định từ này sẽ làm thay đổi những từ thể hiện tình cảm tích cực sang tiêu cực. Như vd 3, người sử dụng nhận xét một sản phẩm trong hai vế, ở vế đầu mang ý nghĩa tích cực, nhưng ở vế sau thì ngược lại, vì vậy hệ thống cho kết quả là trung tính. Tuy nhiên trong vd 4, về mặt ý nghĩa thì ví dụ này mang tính tiêu cực nhưng hệ thống phân tích sai ở vế thứ hai về sự mong muốn tích cực trong tương lai, vì vậy hệ thống đánh giá là trung tính.

Ngoài ra, hệ thống có thể đánh giá bình luận nhiều thực thể. Việc xác định những từ thể hiện tình cảm đối với từng thực thể. Đối với vd 5 như hệ thống phân tích đối với một thực thể thì hệ thống nhận biết được Asus mang hàm ý tiêu cực, còn thực thể Dell Vostro ngược lại mang ý nghĩa tích cực.

## 5. CÔNG TRÌNH LIÊN QUAN

### 5.1. Phân tích những ý kiến dự đoán trên Web

Trong thời đại bùng nổ công nghệ thông tin như hiện nay, việc các công ti hay một tổ chức tự xây dựng cho mình một trang web là rất phổ biến, trang Web này như là một kênh thông tin riêng hoặc thể hiện tiềm năng của họ. Ở đó, họ có thể giới thiệu những gì mà họ đang làm như là các sản phẩm, các dịch vụ, các cuộc thảo luận về vấn đề xã hội, buôn bán, thu thập ý kiến của khách hàng [11].

Trong các bài nghiên cứu [12], [13], [14], [15], [16], [17], [18], các tác giả thường phân tích và khai thác các ý kiến đánh giá của khách hàng trên các sản phẩm như là điện thoại, máy ảnh, máy tính, sách, phim ảnh. Những ý kiến này thường thể hiện là thích hay không thích các sản phẩm đó, người ta gọi những ý kiến này là những ý kiến nhận định hoặc phán xét. Tuy nhiên những ý kiến này không phải là tiếng Việt.

Và trong bài nghiên cứu [11] tác giả đề cập tới một ý kiến khác với những nghiên cứu trên mà tác giả gọi là ý kiến dự đoán. Ta đã nghe khá nhiều về những ý kiến dự đoán về tương lai của một chủ đề như thị trường bất động sản, kết quả của các trận đấu bóng đá hay là các cuộc bầu cử. Những dự đoán này thường dựa trên niềm tin và kiến thức của người đưa ra dự đoán là chính (thường là các chuyên gia). Ví dụ trong câu: "Giá của bất động sản sẽ được giảm xuống trong vài tháng tới", như ta thấy đây là một câu dự đoán ở tương lai về thị trường bất động sản và trong câu này nó cũng thể hiện mặt tích cực trong vế "giá bất động sản sẽ giảm".

Do đó, để đưa ra được các dự đoán về thời tiết, động đất, sóng thần, thì các tác giả cần một cơ sở dữ liệu số (số liệu) rất lớn, tiến hành phân tích trên những con số đó. Tuy nhiên, trong bài [11] không đề cập về những phân tích dự đoán dựa trên việc phân tích số liệu mà tác giả muốn đề cập đến việc phân tích những câu phát biểu dự đoán trong các đoạn văn bản không có cấu trúc xác định dựa trên kỹ thuật xử lý ngôn ngữ tự nhiên (NLP), từ đó đưa ra kết luận về mức độ tích cực của từng dự đoán như ví dụ ở trên.

Cụ thể, trong bài [11] tác giả đã thí nghiệm trên tập các ý kiến dự đoán trên Web của một cuộc bầu cử. Mục đích của tác giả là phân tích tự động tập ý kiến rất lớn từ những người dùng đã bày tỏ trên web, từ đó rút ra được Đảng nào đang chiếm ưu thế về niềm tin của người dân và đưa ra được con số phần trăm thắng cử của mỗi Đảng tham gia vào cuộc bầu cử này. Vì thế tác giả đã đưa ra mô hình đánh giá như sau:

*ElectionPredictionOpinion = (Party, Valence).*

- *ElectionPredictionOpinion:* Kết quả đánh giá cho mỗi Đảng
- *Party***:** là Đảng mà người bình luận đang muốn nói đến.
- *Valence***:** là tỷ lệ phần trăm thắng cử mà hệ thống tính toán được.

Tác giả cũng xây dựng hệ thống này gồm có ba bước chính được huấn luyện bởi học máy sử dụng chức năng n-gram và SVM (Support Vector Machine) để lượng giá:

1. Chức năng tổng quát hóa: chức năng này làm công việc xác định những ai liên quan đến Đảng đang nói đến trong câu thì qui về tên của Đảng mà người đó đang làm việc.

2. Phân loại và đánh giá từng câu ý kiến dự đoán sử dụng kỹ thuật SVM dựa trên mô hình (Party, Valence).

3. Tổng hợp lại tất cả các phân loại mà bước 2 đã làm và đưa ra được kết quả là phần trăm của Đảng được dự đoán thắng cuộc.

Như vậy, ưu điểm của hệ thống trên là việc tổng quát hóa các đối tượng liên quan lại thành một đối tượng để đánh giá, giúp cho việc đánh giá được chính xác hơn.

**5.2. Xây dựng một miền Ontology tự động từ một mạng ngữ nghĩa**

Khai thác ý kiến và phân tích tình cảm là nhánh con của NLP và khai thác văn bản (text mining) nhằm mục đích khám phá và khai thác tự động tri thức về tình cảm con người, sự đánh giá những ý kiến từ những dữ liệu văn bản gốc như là trang nhật ký của một người nào đó, những nhận xét trên website và trong những bản phản hồi từ khách hàng.

Những tác giả trong công trình [19] đã đề xuất một phương pháp xây dựng một miền Ontology tự động từ một mạng ngữ nghĩa ConceptNet. ConceptNet này được xây dựng bởi các tình nguyện viên trên thế giới. Và kết quả của công trình trên có thể sử dụng như nguồn từ vựng hoặc thực hiện xác định các mục tiêu để phân tích tình cảm trong thời gian đó.

Khác với những công trình trước (được đề cập đến trong 5.1), trong công trình nghiên cứu [19] các tác giả đã đưa ra được giải pháp cải tiến hơn so với những giải pháp truyền thống. Sự khác biệt trong giải pháp của tác giả so với những giải pháp khác là tác giả đã đề xuất một mạng ngữ nghĩa của common-sense knowledge-base (ConceptNet) để tạo tự động ra một miền ontology của những tính năng của sản phẩm và các thuộc tính của nó. Những công trình trước thường đánh giá chung những danh sách tính năng của sản phẩm, trong khi đó tác giả tạo ra một ontology ở đó những tính năng của sản phẩm như một khái niệm hoặc là những nút trên mạng ngữ nghĩa được kết nối với những nút khác sử dụng nhiều kiểu của quan hệ ngữ nghĩa (semantic relationship - theo nhiều kiểu khác nhau hoặc nhiều mối quan hệ khác nhau). Vì thế mà sản phẩm sẽ xuất hiện trong miền ontology và từ vựng đã tạo ra từ cách này mang ngữ nghĩa phong phú hơn là từ vựng.

Tác giả tận dụng ConceptNet để xây dựng miền ontology. Các nút thể hiện các khái niệm, các cạnh thể hiện các tính chất, thuộc tính, quan hệ. Tác giả không những đưa ra những mối quan hệ ngữ nghĩa giống như IsA, HasA mà còn nhiều mối quan hệ khác CreatedBy, MadeOf, PartOf, DesireOf và DefineAs. Phạm vi của ConceptNet là những tri thức chung và không giới hạn bởi một miền cụ thể nào, rất là hữu dụng trong việc khai thác những câu bình luận về những tính năng của sản phẩm.

Bên cạnh đó, các công trình [20], [21] và [22] có sử dụng ontology về thực thể có tên nhưng là để tăng hiệu quả truy hồi tài liệu. Còn công trình [23] là nhận diện nhận xét rác chứ không phải nhận diện quan điểm của nhận xét.

# 6. KẾT LUẬN

Chúng tôi đã đề xuất phương pháp sử dụng Ontology kết hợp từ điển để xác định mức độ thể hiện tình cảm trong nhận xét của khách hàng. Trong phương pháp này, chúng tôi đã xác định được những thực thể xuất hiện trong miền Ontology. Sau đó, dựa vào tập từ điển chúng tôi đã xác định các từ thể hiện tình cảm và điểm số của chúng (tính từ, động từ, trợ từ, phủ định từ). Từ đó, chúng tôi tiến hành đánh giá điểm số, tổng hợp và đưa ra được kết luận tích cực, tiêu cực hay trung tính cho từng thực thể tương ứng. Qua kết quả thí nghiệm trên ba tập dữ liệu gồm 200 câu về một thực thể chủ đề điện thoại di động, 100 câu về nhiều thực thể chủ đề điện thoại di động, máy tính xách tay, và 100 câu nhiều thực thể kết hợp cả hai chủ đề trên, phương pháp đánh giá trên tập một thực thể ở mỗi bình luận cao hơn hai tập dữ liệu còn lại dựa trên đánh giá điểm số cho từng thực thể tương ứng ở cả hai độ đo được đề cập ở trên.

Hệ thống của chúng tôi chủ yếu tập trung vào xác định và đánh giá những ý kiến của khách hàng thông qua hai loại sản phẩm chính đó là điện thoại di động và máy tính xách tay, đây là những loại mặt hàng đang phổ biến hiện nay. Trong tương lai, nếu có điều kiện chúng tôi muốn mở rộng ra nhiều chủ đề khác như các bình luận trên các bài báo điện tử hoặc về những bình luận trên các sản phẩm khác.

Ngoài ra, trong hệ thống của chúng tôi chỉ đánh giá cho các bình luận thuộc thể loại câu khẳng định và so sánh, mà chưa đề cập đến những mẫu câu khác như câu nhân hóa, câu điều kiện và câu cảm thán. Vì thế đây là một vấn đề cần nghiên cứu về sau để xác định được quan điểm đúng của các câu thuộc các thể loại này.

## TÀI LIỆU THAM KHẢO

1. Scholz, T. and Conrad, S. - Opinion Mining in Newspaper Articles by Entropy-Based Word Connections. In Proceedings of the 2013 Conference on Empirical Methods in Natural Language (EMNLP-2013), ACL (2013)1828-1839.

2. Feldman, R. - Techniques and Applications for Sentiment Analysis. In Communications of the ACM, **56** (2013) 82-89.

3. Bakliwal, A. et al. - Mining Sentiments from Tweets. In Proceedings of the 3rd Workshop on Computational Approaches to Subjectivity and Sentiment Analysis, ACL (2012) 11–18.

4. Bakliwal, A. et al. - Sentiment Analysis of Political Tweets: Towards an Accurate Classifier. In Proceedings of the Workshop on Language in Social Media (LASM-2013), ACL (2013) 49–58.

5. Korayem, M., Crandall, D. and Abdul-Mageed, M. - Subjectivity and Sentiment Analysis of Arabic: A Survey. In Proceedings of the first International Conference on Advanced Machine Learning Technologies and Applications (AMLTA12), Springer, LNCS 322 (2012) 128-139.

6. Abdul-Mageed, M., Diab, M. and Korayem, M. - Subjectivity and Sentiment Analysis of Modern Standard Arabic. In Proceedings of the 49th Annual Meeting of the Association for Computational Linguistics: Human Language Technologies: short papers, ACL (2011) 587-591.

7. Duong, T. H. H., Vu, T. D. and Ngo, V. M. - *Detecting Vietnamese Opinion Spam*. In Proceedings of Scientific Researches on the Information and Communication Technology in 2012 (ICTFIT'12), Publishers of Engineering Sciences, Vietnam (Vietnamese title: "*Phát hiện đánh giá spam cho tiếng Việt*"), (2012) 53-59.

8. Fensel, D., Harmelen V. F. and Horrocks, I. - OIL: An Ontology Infrastructure for the Semantic Web. In IEEE Intelligent System, 16(2) (2001) 38-45.

9. Ding, L. et al. - Using Ontologies in the Semantic Web: A Survey. Book Chapter in Sharman, R., ed al.: Ontologies - A Handbook of Principles, Concepts and Applications in Information Systems, Book of series Integrated Series in Information Systems, 14(I) (2007) 77-113.

10. Marsh, E. and Perzanowski, D. - MUC-7 Evaluation of IE Technology: Overview of Results. In Proceeding of the Seventh Message Understanding Conference (MUC-7) (1998).

11. Kim S. and Eduard H. - Crystal: Analyzing Predictive Opinions on the Web. In Joint Conference on Empirical Methods in Natural Language Processing and Computational Natural Language Learning (EMNLP – CoNLL) (2007) 1056-1064.

12. Hu, M. and Liu, B. - Mining and Summarizing Customer Reviews. In Proceedings of 10th ACM SIGKDD International Conference on Knowledge Discovery and Data Mining (KDD) (2004) 166-177.
13. Hu, M. and Liu, B. - Mining Opinion Features in Customer Reviews. In Proceedings of 19th National Conference on Artificial Intelligence(AAAI) (2004) 755-761.
14. Alexander O. - Sentiment Mining for Natural Language Documents. In COMP3006 PROJECT REPORT, Computer Science Research Project, Department of Computer Science Australian National University (2009).
15. Casey W., Navendu G. and Shlomo A. - Using Appraisal Groups for Sentiment Analysis. In Proceedings of the 14th ACM International Conference on Information and Knowledge Management (CIKM) (2005) 625-631.
16. Ramanathan, N. Bing, L. and Alok, C. - Sentiment Analysis of Conditional Sentences. In Proceedings of 2009 Conference on Empirical Methods in Natural Language Processing (CEMNLP) (2009) 180-189.
17. Khin, S. - Ontology Based Combined Approach for Sentiment Classification. In Proceedings of 3th International Conference on Communications and Information Technology (CIT) (2009) 112-115.
18. Ginsca L., et al. - Sentimatrix – Multilingual Sentiment Analysis Service. In Proceedings of 2nd Workshop on Computational Approaches to Subjectivity and Sentiment Analysis (WASSA), ACL (2011) 189-195.
19. Ashish S., Vikram G., Denzil C. and Anirban M. - Generating Domain-Specific Ontology from Common-Sense Semantic Network for Target-Specific Sentiment Analysis. In Proceedings of 5th International Conference (GWC) of the Global WordNet Association (GWA) (2010).
20. Ngo, V. M., Cao, T.H. Discovering Latent Concepts and Exploiting Ontological Features for Semantic Text Search. In Proceedings of the 5th International Joint Conference on Natural Language Processing (IJCNLP-2011), (2011) 571-579.
21. Cao, T. H., Ngo, V. M. Semantic Search by Latent Ontological Features. In International Journal of New Generation Computing, Springer-Verlag, SCI, Vol. 30, No.1, (2012) 53-71
22. Ngo, V. M. Discovering Latent Information by Spreading Activation Algorithm for Document Retrieval. In International Journal of Artificial Intelligence & Applications, Vol. 5, No. 1, (2014) 23-34
23. Nguyen, L. H., Pham, N. T. H., Ngo, V. M. Opinion Spam Regconition Method for Online Reviews using Ontological Features. In Journal of Science, Special Issue: Natural Science and Technology, Ho Chi Minh City University of Education, Vol. 61(95), (2014) 44-59.